\documentclass[11pt]{article}

\usepackage[final]{acl}

\usepackage{times}
\usepackage{latexsym}

\usepackage[T1]{fontenc}

\usepackage[utf8]{inputenc}

\usepackage{microtype}

\usepackage{inconsolata}

\usepackage{graphicx}

\usepackage{listings}
\usepackage{multicol}
\usepackage{multirow}
\usepackage{booktabs}
\usepackage{amsmath}
\usepackage{enumitem}
\usepackage{subcaption}

\usepackage{tcolorbox}
\usepackage{listings}
\lstdefinestyle{plain}{
    basicstyle=\fontsize{7}{9.5}\ttfamily,
    keywordstyle=\color{blue},
    commentstyle=\color{gray},
    stringstyle=\color{green},
    showstringspaces=false,
    breaklines=true,
    breakatwhitespace=false,
    breakindent=0pt,
    escapeinside={(*@}{@*)}
}

\definecolor{MutedGreen}{RGB}{85, 170, 85}
\definecolor{CoolAccent}{RGB}{120, 145, 230}
\definecolor{IAP}{RGB}{255, 126, 121}
\definecolor{CoT}{RGB}{91, 155, 213}
\definecolor{WarmOrange}{RGB}{255, 165, 85}

\title{SAFARI: An Industrial Benchmark for LLM-Assisted Hazard Analysis and Risk Assessment}

\author{Chenxi Wu\thanks{Equal contribution.}, Zimu Wang$^*$, Haiyang Zhang\thanks{Corresponding author.}, Wei Wang, Zhijie Xu \\
  School of Advanced Technology, Xi'an Jiaotong-Liverpool University \\
  \texttt{\{Chenxi.Wu25, Zimu.Wang19\}@student.xjtlu.edu.cn} \\
  \texttt{\{Haiyang.Zhang, Wei.Wang03, Zhijie.Xu\}@xjtlu.edu.cn}
  \\}

\begin{document}
\maketitle
\begin{abstract}
Large language models (LLMs) are increasingly considered for safety-critical engineering, yet their reliability in regulated functional-safety workflows remains underexplored. We introduce \textbf{SAFARI} (\textbf{S}afety-\textbf{A}ware \textbf{F}unctional \textbf{A}utomotive \textbf{R}isk \textbf{I}nference), the first industrial benchmark for LLM-assisted automotive Hazard Analysis and Risk Assessment (HARA) under ISO 26262. It contains 3,000 de-identified industrial HARA cases and evaluates two coupled tasks: \textit{open-ended hazard analysis} and \textit{standards-grounded risk assessment}. To evaluate open-ended HARA artifacts, we propose the first reference-anchored LLM-as-a-judge protocol with high expert correlation. Experiments with nine frontier LLMs show that models often produce plausible hazard narratives but remain weak at ISO 26262 risk classification, with the best ASIL macro-F1 reaching only 0.261. Chain-of-Thought prompting provides limited benefit and often degrades categorical risk assessment. Error analysis further localizes major failures to scenario-critical context omissions during hazard generation and to controllability misjudgments during risk assessment, indicating where expert oversight should be concentrated. The dataset can be obtained from \url{https://github.com/xixi47520-hash/HARA}.
\end{abstract}

\section{Introduction}

Modern vehicles increasingly rely on software-defined and distributed control functions across perception, planning, control, and actuation \citep{10136910,10156892}. Consequently, safety assessment
must reason not only about component failures, but also about their interaction with operational scenarios and system-level behavior \citep{leveson2016engineering}. Under ISO~26262, Hazard Analysis and Risk Assessment (HARA) is a core concept-phase activity in which engineers identify hazardous events caused by malfunctions, assign \textit{Severity} (S), \textit{Exposure} (E), and \textit{Controllability} (C), and derive the Automotive Safety Integrity Level (ASIL) that governs downstream safety requirements. HARA therefore combines open-ended hazard reasoning with standards-governed risk classification: ASIL is deterministically derived from S/E/C, while the underlying risk parameters require expert judgment about traffic situations, vehicle dynamics, cause propagation, and human controllability. A locally plausible hazard description may still lead to an incorrect controllability rating or downstream ASIL allocation, making HARA a high-stakes, cascade-sensitive reasoning workflow.

Large language models (LLMs) offer strong potential to assist HARA because of their capabilities to reason and generate structured technical text with step-by-step rationales~\citep{wei2023chainofthoughtpromptingelicitsreasoning,10.1609/aaai.v39i23.34607,kang-etal-2025-grpo,KANG2026113355,zhao2026audioprocessbench}. However, fluency is not enough for industrial safety engineering, as safety evaluations necessitate evidence on which HARA subtasks LLMs can support under expert oversight, which outputs remain unreliable, and where human review should be concentrated. Existing evaluations do not adequately provide this evidence. General LLM safety benchmarks primarily examine truthfulness, harmful outputs, and refusal behaviors~\citep{guo-etal-2025-lost,shen2026psychethicsbenchevaluatinglargelanguage,xu2026harmexposinghiddenvulnerabilities}, while domain-specific benchmarks are dominated by finance, medicine, and law~\citep{guha2023legalbenchcollaborativelybuiltbenchmark,chen-etal-2025-medfact,peng2026herculeanagenticbenchmarkfinancial}. However, the evaluation of HARA remains underexplored, as it captures the combination of open-ended causal hazard generation, standards-based S/E/C classification, and deterministic ASIL propagation. Therefore, the construction of an industrial-scale, model-comparative benchmark that localizes failures across the HARA workflow is worthwhile.

\begin{figure}[t]
    \centering
    \includegraphics[width=\linewidth]{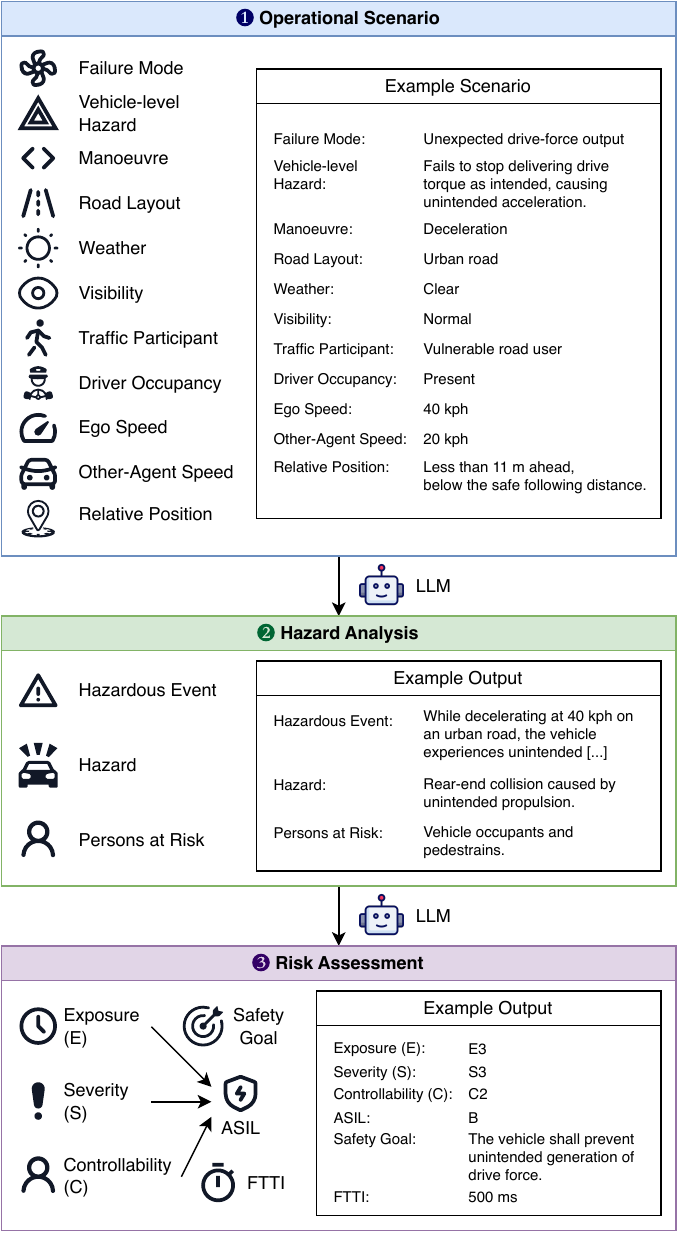}
    \caption{Overall workflow of the SAFARI benchmark, which combines hazard analysis and risk assessment within a single HARA workflow.}
    \label{fig:SAFARI:-overview}
\end{figure}

We present \textbf{SAFARI} (\textbf{S}afety-\textbf{A}ware \textbf{F}unctional \textbf{A}utomotive \textbf{R}isk \textbf{I}nference), a large-scale industrial benchmark for evaluating LLMs on automotive HARA under ISO~26262. SAFARI contains 3{,}000 de-identified industrial HARA cases centered on high-criticality unintended drive force or torque output.
To reflect the structure of practical HARA workflows, SAFARI is divided into two sequential tasks, as shown in Figure \ref{fig:SAFARI:-overview}. \textbf{Task~1} evaluates hazard analysis, requiring models to generate the \textit{hazardous event}, \textit{hazard}, and \textit{persons at risk} from a structured operational scenario. \textbf{Task~2} evaluates risk assessment, requiring models to infer \textit{Exposure} (E), \textit{Severity} (S), \textit{Controllability} (C), the \textit{derived ASIL}, \textit{Safety Goal}, and \textit{Fault Tolerant Time Interval} (FTTI). This decomposition separates open-ended hazard reasoning from standards-constrained risk classification while preserving their causal dependency within the HARA process.
By focusing on a single safety-critical malfunction class across diverse scenarios, our benchmark enables controlled evaluation while preserving industrial realism.

We conduct experiments on nine frontier LLMs under both few-shot and chain-of-thought (CoT) prompting. To evaluate open-ended HARA artifacts whose correctness depends on scenario-specific causal chains and safety intent, we introduce the first HARA-specific, reference-anchored LLM-as-a-judge protocol, which scores outputs along four safety-specific dimensions: \textit{hazard-chain consistency}, \textit{physical-consequence consistency}, \textit{risk-population coverage}, and \textit{safety-goal alignment}, and shows strong agreement with certified engineers.
Experimental results reveal a clear capability boundary between hazard analysis and standards-grounded risk assessment. While frontier LLMs generally perform well on hazardous-event generation and risk-population identification, performance drops substantially on E/S/C prediction and downstream ASIL allocation. We further find that CoT prompting does not consistently improve HARA reasoning and often degrades categorical risk assessment, although it benefits FTTI estimation. Failure localization identifies two major bottlenecks: contextual hazard interpretation and controllability assessment, both of which contribute to downstream risk-classification errors.

The key contributions of this paper are summarized as follows:

\begin{itemize}[leftmargin=*,nosep]
    \item We introduce SAFARI, the first industrial benchmark for evaluating LLM-assisted HARA under ISO~26262, containing 3{,}000 de-identified cases across hazard analysis and risk assessment.
    \item We propose a unified evaluation framework that combines automatic metrics for structured risk labels with a reference-anchored, expert-validated LLM-as-a-judge protocol for open-ended safety artifacts.
    \item We benchmark nine frontier LLMs and reveal a clear gap between plausible hazard generation and risk assessment, and further localize major failures to scenario-context omission and controllability misjudgment.
\end{itemize}

\section{Background and Related Work}

\paragraph{HARA and Functional Safety.}

ISO~26262\footnote{\url{https://www.iso.org/standard/68383.html}} defines HARA as the concept-phase process used to identify hazardous events, assess \textit{Exposure} (E), \textit{Severity} (S), and \textit{Controllability} (C), and \textit{derive ASILs}.
Prior work has improved the structure and traceability of HARA through scenario terminology and representation \citep{7313256,8500406}, ontology-based scene modeling \citep{8500632}, and hazardous-event identification \citep{7535462,9304780}. While these studies provide foundations for representing operational contexts, deriving hazardous events, and improving traceability, they primarily address how HARA artifacts are modeled or generated for specific systems, while the core assessment of E/S/C and ASIL remains dependent on expert judgment. SAFARI builds on this line of work by formulating HARA as an empirical evaluation problem over real industrial HARA artifacts, covering both hazardous-event formulation and standards-grounded risk assessment.

\paragraph{LLMs for Safety-Critical Engineering.}

NLP techniques have been applied to safety-related artifacts, including harmful-content detection, ethics evaluation, and vulnerability probing \citep{10580402,guo-etal-2025-lost,shen2026psychethicsbenchevaluatinglargelanguage}. However, such work primarily evaluates general language model safety rather than standards-governed engineering analysis. In safety-critical engineering, LLMs have been explored as assistants for safety-critical engineering, including hazard analysis, requirements engineering, STPA-based safety assessment, and automotive functional-safety workflows \citep{10564681,nouri2024engineeringsafetyrequirementsautonomous,shi-etal-2024-aegis,QI2025100622}. While these works demonstrate the potential of LLMs to reduce manual effort and support expert review, they are primarily case-study- or system-oriented and do not provide a large-scale benchmark for evaluating HARA reasoning. Different from these studies, SAFARI evaluates multiple frontier LLMs on real industrial HARA artifacts, covering both open-ended hazard reasoning and standards-grounded E/S/C assessment, ASIL allocation, safety goal generation, and FTTI estimation.

\section{Dataset Construction}
\label{sec:dataset}

\subsection{Scope and Source}
\label{sec:dataset:scope}

SAFARI is constructed from de-identified HARA artifacts collected from industrial automotive safety-engineering projects. To enable controlled evaluation, we focus on a safety-critical powertrain malfunction---\emph{unintended drive force or torque output}---and vary the operational context along ISO~26262-relevant factors, including manoeuvre, road layout, weather, visibility, traffic participants, driver occupancy, ego and other-agent speeds, and relative position.
This design fixes the failure mechanism while varying the conditions that determine Exposure, Severity, and Controllability. SAFARI therefore isolates scenario-dependent risk reasoning from the simpler problem of recognizing the failure mode.

The source artifacts were produced in structured HARA workshops across multiple industrial projects by certified functional-safety engineers. Each benchmark instance contains \textbf{(1)} a parametric operational context, \textbf{(2)} expert reference texts for the hazardous event, hazard, and persons at risk, and \textbf{(3)} expert annotations for Exposure, Severity, Controllability, derived ASIL, safety goal, and FTTI. Before benchmark construction, all samples were scrubbed of vehicle identifiers, project names, supplier identifiers, and fine-grained geographic information.

\subsection{Schema}
\label{sec:dataset:schema}

\begin{table}[t]
\centering
\small
\setlength{\tabcolsep}{4mm}
\begin{tabular}{l|ll}
\toprule
\textbf{Block} & \textbf{Field} & \textbf{Type} \\
\midrule
\multirow{11}{*}{Input}
  & Failure Mode              & Categorical \\
  & Vehicle-level Hazard      & Categorical \\
  & Manoeuvre                 & Categorical \\
  & Road Layout               & Categorical \\
  & Weather                   & Categorical \\
  & Visibility                & Categorical \\
  & Traffic Participant       & Categorical \\
  & Driver Occupancy          & Categorical \\
  & Ego Speed                 & Numerical \\
  & Other-Agent Speed         & Numerical \\
  & Relative Position         & Textual \\
\midrule
\multirow{3}{*}{Task~1}
  & Hazardous Event           & Textual \\
  & Hazard                    & Textual \\
  & Persons at Risk           & Textual \\
\midrule
\multirow{6}{*}{Task~2}
  & Exposure (E)              & Ordinal \\
  & Severity (S)              & Ordinal \\
  & Controllability (C)       & Ordinal \\
  & ASIL                      & Derived \\
  & Safety Goal               & Textual \\
  & FTTI                      & Numerical \\
\bottomrule
\end{tabular}
\caption{Compact schema of the SAFARI benchmark.}
\label{tab:schema}
\end{table}

Each sample contains eleven input fields and nine output fields, summarized in Table~\ref{tab:schema}. The input fields specify a fixed malfunction context and a parameterized operational scenario. In particular, \textit{Failure Mode} and \textit{Vehicle-level Hazard} serve as constant anchors for the studied malfunction, while the remaining fields describe scenario factors that affect ISO~26262 risk assessment, such as manoeuvre, road layout, environmental conditions, traffic participants, vehicle speeds, and relative position. We retain speed values in their original unit-bearing textual form (e.g., \texttt{10 kph}) to preserve the representation used in industrial HARA artifacts.

The output fields are organized according to the two-stage HARA workflow. Task~1 contains open-ended textual artifacts: \textit{Hazardous Event}, \textit{Hazard}, and \textit{Persons at Risk}. These fields evaluate whether a model can transform a structured scenario into a safety-relevant causal description. Task~2 contains the downstream risk-assessment outputs: ordinal ISO~26262 risk parameters \textit{Exposure}, \textit{Severity}, and \textit{Controllability}, the \textit{derived ASIL}, and additional safety-engineering artifacts including \textit{Safety Goal} and \textit{FTTI}. ASIL is not treated as an independent free-form label; it is deterministically derived from the E/S/C tuple using the ISO~26262 lookup table. This schema, therefore, captures both open-ended hazard formulation and standards-constrained risk reasoning within a single benchmark instance. A representative example is shown in Figure~\ref{fig:SAFARI:-overview}.

\subsection{Annotation and Quality Assurance}
\label{sec:dataset:quality}

Our SAFARI benchmark is derived from HARA artifacts originally produced for industrial automotive safety engineering projects, rather than from crowdsourced or post-hoc annotation. The source analyses were conducted in structured HARA workshops by certified functional-safety engineers with experience in ISO~26262-compliant safety analysis, and underwent internal peer review, expert evaluation, and standard project quality-assurance procedures before benchmark construction.

\begin{figure}[t!]
    \centering
    \begin{subfigure}[b]{0.49\linewidth}
        \includegraphics[width=\textwidth]{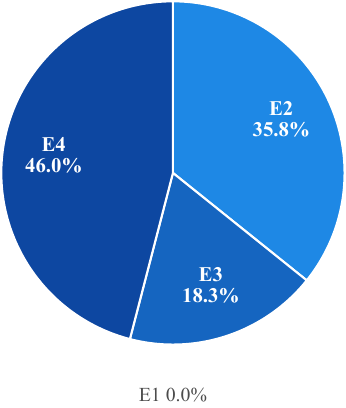}
        \caption{Exposure (E)}
        \label{fig:exposure}
    \end{subfigure}
    \hfill
    \begin{subfigure}[b]{0.49\linewidth}
        \includegraphics[width=\textwidth]{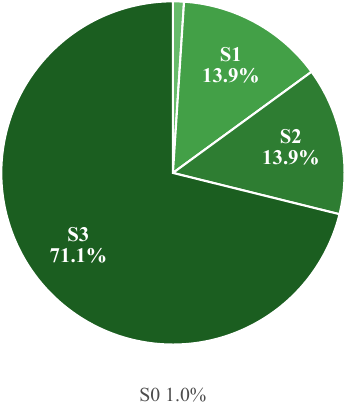}
        \caption{Severity (S)}
        \label{fig:severity}
    \end{subfigure}
    \begin{subfigure}[b]{0.49\linewidth}
        \includegraphics[width=\textwidth]{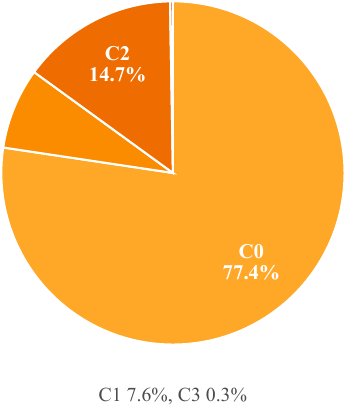}
        \caption{Controllability (C)}
        \label{fig:controllability}
    \end{subfigure}
    \hfill
    \begin{subfigure}[b]{0.49\linewidth}
        \includegraphics[width=\textwidth]{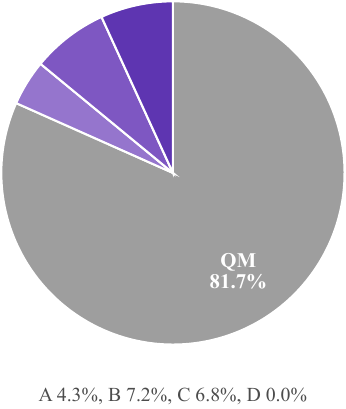}
        \caption{Derived ASIL}
        \label{fig:asil}
    \end{subfigure}
    \caption{Label distributions over SAFARI samples.}
    \label{fig:label-distribution}
\end{figure}

\begin{figure}[t!]
    \centering
    \begin{subfigure}[b]{0.49\linewidth}
        \includegraphics[width=\textwidth]{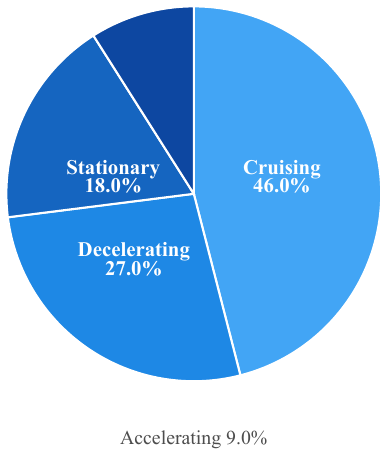}
        \caption{Manoeuvre}
        \label{fig:manoeuvre-distribution}
    \end{subfigure}
    \hfill
    \begin{subfigure}[b]{0.49\linewidth}
        \includegraphics[width=\textwidth]{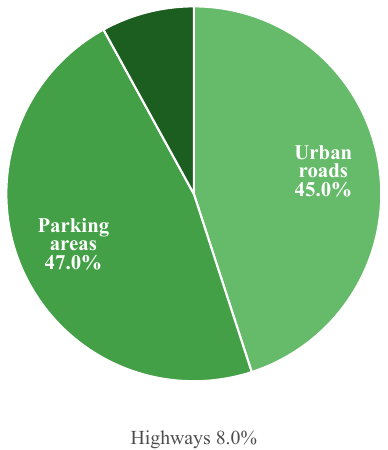}
        \caption{Road Layout}
        \label{fig:road-layout-distribution}
    \end{subfigure}

    \begin{subfigure}[b]{0.49\linewidth}
        \includegraphics[width=\textwidth]{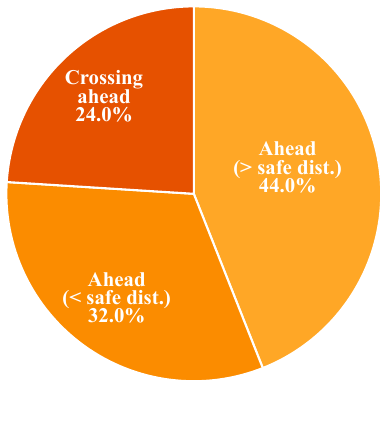}
        \caption{Relative Position}
        \label{fig:relative-position-distribution}
    \end{subfigure}
    \hfill
    \begin{subfigure}[b]{0.49\linewidth}
        \includegraphics[width=\textwidth]{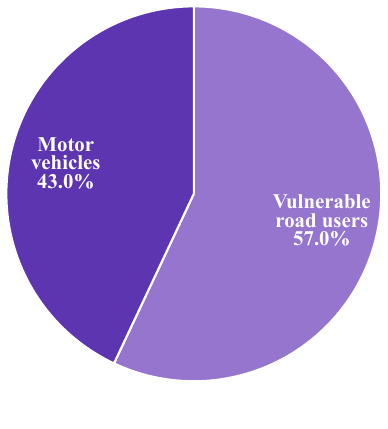}
        \caption{Traffic Participant}
        \label{fig:traffic-participant-distribution}
    \end{subfigure}
    \caption{Attribute distributions over SAFARI samples.}
    \label{fig:scenario-attribute-distribution}
\end{figure}

For benchmark annotation, all annotators underwent a calibration phase before formal annotations, which covers the SAFARI schema, ISO 26262 risk-parameter definitions, ASIL derivation rules, and representative edge cases for hazardous-event formulation, controllability assessment, safety goal writing, and FTTI estimation. After calibration, each sample was independently reviewed by two functional-safety engineers. The review covered both Task~1 textual references---\textit{Hazardous Event}, \textit{Hazard}, and \textit{Persons at Risk}---and Task~2 risk-assessment annotations, including \textit{Exposure}, \textit{Severity}, \textit{Controllability}, \textit{Safety Goal}, and \textit{FTTI}. Before adjudication, the two reviewers achieved approximately 90\% agreement on the reviewed annotations. Remaining disagreements were resolved by a senior functional-safety engineer.
After adjudication, all structured risk labels were checked against ISO~26262-3 consistency rules, including verification that each $(E,S,C)$ tuple maps to the correct derived ASIL. As a result, SAFARI contains calibrated, expert-reviewed, open-ended references and consistency-verified, structured risk labels.

\subsection{Dataset Characteristics}
\label{sec:dataset:characteristics}

\begin{table}[t]
\centering
\small
\begin{tabular}{lccc}
\toprule
\textbf{Dimension} &
\textbf{Spearman $\rho$} &
\textbf{Pearson $r$} &
\textbf{Kendall $\tau$}\\
\midrule
HLM & 0.807 & 0.890 & 0.876 \\
PCC & 0.892 & 0.895 & 0.870 \\
RCC & 0.941 & 0.843 & 0.832 \\
SGA & 0.951 & 0.953 & 0.934 \\
\bottomrule
\end{tabular}
\caption{Judge-expert agreement on the validation split.}
\label{tab:judge-agreement}
\end{table}

The final SAFARI dataset contains 3{,}000 samples, split into 1{,}600 training, 400 validation, and 1{,}000 test instances, with stratification by scenario and ASIL-relevant fields. Figures~\ref{fig:label-distribution} and \ref{fig:scenario-attribute-distribution} summarize the distributions of labels and operational attributes over all 3{,}000 samples. As expected from industrial HARA workflows, SAFARI exhibits a long-tailed distribution across multiple risk categories. We intentionally preserve this empirical distribution since rare but safety-critical situations are inherently less frequent in practice. SAFARI therefore reflects the conditions encountered in real-world functional-safety engineering.

\section{Experiments}
\label{sec:evaluation}

\subsection{Evaluation Metrics}

Given that SAFARI contains both structured risk labels and open-ended safety artifacts, we use complementary automatic and reference-anchored metrics for comprehensive evaluation.

\paragraph{Automatic Metrics.}
For structured risk assessment, Exposure (E), Severity (S), Controllability (C), and derived ASIL are evaluated using \textit{accuracy} and \textit{macro-F1}, in which macro-F1 is reported to account for the imbalanced empirical label distribution. FTTI is evaluated using mean absolute error (MAE). When expert annotations specify inequality constraints, predictions satisfying the constraint incur zero error; otherwise, error is computed with respect to the nearest constraint boundary.

\paragraph{Reference-anchored LLM-as-a-Judge.}
Open-ended HARA artifacts cannot be reliably evaluated by surface-form matching, since multiple textual realizations may be semantically valid while differing in wording. We therefore introduce a reference-anchored LLM-as-a-judge protocol for Task~1 outputs and safety goal evaluation. For each sample, the judge receives the structured scenario, the expert reference, and the model prediction, and assigns an ordinal score of 1--5 according to a dimension-specific rubric. Scores are normalized to $[0,1]$ for reporting, with higher values indicating stronger semantic agreements to reference.

We prompt the judge to ground in four specific dimensions: \textit{Hazard Logic Mapping} (HLM) for hazardous-event causal chain and scenario grounding, \textit{Physical Consequence Consistency} (PCC) for hazard-level consequence consistency, \textit{Risk Coverage Completeness} (RCC) for coverage of persons at risk, and \textit{Safety Goal Alignment} (SGA) for preservation of the intended safety objective. Qwen3.6-Max is used as the judge for evaluation. We sample 180 outputs and compare judge scores with ratings from certified ISO~26262 functional-safety engineers and report Spearman's $\rho$, Pearson's $r$, and Kendall's $\tau$ in Table \ref{tab:judge-agreement}.

\begin{table}[t]
\centering
\small
\setlength{\tabcolsep}{2.2pt}
\begin{tabular}{l|c|cccc}
\toprule
\textbf{Model} & \textbf{Strategy} & \textbf{HLM} & \textbf{PCC} & \textbf{RCC} & \textbf{Avg.} \\
\midrule
\multirow{2}{*}{\href{https://openai.com/index/introducing-gpt-5-5/}{GPT-5.5}} & FS  & \textbf{0.200} & 0.781 & 0.978 & 0.653 \\
 & CoT & 0.185 & 0.539 & 0.940 & 0.555 \\
\midrule
\multirow{2}{*}{\href{https://openai.com/index/introducing-gpt-5-4-mini-and-nano/}{GPT-5.4-mini}} & FS  & 0.197 & 0.626 & 0.971 & 0.598 \\
 & CoT & 0.178 & 0.535 & 0.976 & 0.563 \\
\midrule
\multirow{2}{*}{\href{https://deepmind.google/models/gemini/flash/}{Gemini-3.5-Flash}} & FS  & 0.192 & 0.985 & 0.970 & 0.715 \\
 & CoT & 0.185 & 0.485 & 0.680 & 0.450 \\
\midrule
\multirow{2}{*}{\href{https://api-docs.deepseek.com/news/news260424}{DeepSeek-V4-Pro}} & FS  & 0.192 & 0.874 & 0.989 & 0.685 \\
 & CoT & 0.186 & 0.469 & 0.862 & 0.506 \\
\midrule
\multirow{2}{*}{\href{https://api-docs.deepseek.com/news/news260424}{DeepSeek-V4-Flash}} & FS  & 0.197 & 0.945 & 0.999 & 0.714 \\
 & CoT & 0.186 & 0.474 & 0.946 & 0.535 \\
\midrule
\multirow{2}{*}{\href{https://qwen.ai/blog?id=qwen3.5}{Qwen3.5-397B-A17B}} & FS  & 0.186 & 0.932 & 0.956 & 0.691 \\
 & CoT & 0.185 & 0.372 & 0.946 & 0.501 \\
\midrule
\multirow{2}{*}{\href{https://qwen.ai/blog?id=qwen3.5}{Qwen3.5-122B-A10B}} & FS  & 0.185 & 0.788 & 0.996 & 0.656 \\
 & CoT & 0.183 & 0.483 & \textbf{1.000} & 0.555 \\
\midrule
\multirow{2}{*}{\href{https://docs.z.ai/guides/llm/glm-5.1}{GLM-5.1}} & FS  & 0.198 & \textbf{0.988} & 0.993 & \textbf{0.726} \\
 & CoT & 0.192 & 0.511 & 0.977 & 0.560 \\
\midrule
\multirow{2}{*}{\href{https://www.minimax.io/models/text/m27}{MiniMax-M2.7}} & FS  & 0.195 & 0.899 & 0.977 & 0.690 \\
 & CoT & 0.195 & 0.533 & 0.935 & 0.554 \\
\bottomrule
\end{tabular}
\caption{Task 1 evaluation results, where the best performance is in \textbf{bold}.}
\label{tab:main-task1}
\end{table}

\begin{table*}[t]
\centering
\small
\begin{tabular}{l|c|cccc|c|c|c}
\toprule
\multirow{2}{*}{\textbf{Model}} & \multirow{2}{*}{\textbf{Strategy}} & \multicolumn{4}{c}{\textbf{Macro-F1}} & \textbf{Acc.} & \multirow{2}{*}{\textbf{SGA} ($\uparrow$)} & \multirow{2}{*}{\textbf{FTTI} ($\downarrow$)} \\
& & \textbf{E} ($\uparrow$) & \textbf{S} ($\uparrow$) & \textbf{C} ($\uparrow$) & \textbf{ASIL} ($\uparrow$) & \textbf{ASIL} ($\uparrow$) & & \\
\midrule
\multirow{2}{*}{\href{https://openai.com/index/introducing-gpt-5-5/}{GPT-5.5}} & FS  & 0.214 & \textbf{0.514} & \textbf{0.216} & \textbf{0.261} & 0.415 & 0.488 & 494.9 \\
 & CoT & 0.131 & 0.245 & 0.058 & 0.122 & 0.214 & 0.304 & 528.2 \\
\midrule
\multirow{2}{*}{\href{https://openai.com/index/introducing-gpt-5-4-mini-and-nano/}{GPT-5.4-mini}} & FS  & 0.227 & 0.398 & 0.172 & 0.217 & 0.312 & 0.399 & 522.2 \\
 & CoT & 0.313 & 0.187 & 0.064 & 0.178 & 0.377 & 0.433 & 436.8 \\
\midrule
\multirow{2}{*}{\href{https://deepmind.google/models/gemini/flash/}{Gemini-3.5-Flash}} & FS  & 0.210 & 0.456 & 0.171 & 0.238 & 0.359 & 0.447 & 455.4 \\
 & CoT & 0.187 & 0.298 & 0.035 & 0.155 & 0.213 & 0.311 & 422.6 \\
\midrule
\multirow{2}{*}{\href{https://api-docs.deepseek.com/news/news260424}{DeepSeek-V4-Pro}} & FS  & 0.210 & 0.393 & 0.127 & 0.208 & 0.253 & 0.341 & 638.9 \\
 & CoT & 0.238 & 0.283 & 0.082 & 0.187 & 0.358 & 0.438 & 439.4 \\
\midrule
\multirow{2}{*}{\href{https://api-docs.deepseek.com/news/news260424}{DeepSeek-V4-Flash}} & FS  & 0.214 & 0.405 & 0.169 & 0.259 & 0.387 & 0.460 & 447.4 \\
 & CoT & 0.206 & 0.237 & 0.067 & 0.202 & 0.446 & 0.497 & 344.2 \\
\midrule
\multirow{2}{*}{\href{https://qwen.ai/blog?id=qwen3.5}{Qwen3.5-397B-A17B}} & FS  & \textbf{0.318} & 0.447 & 0.087 & 0.164 & 0.174 & 0.244 & 605.6 \\
 & CoT & 0.288 & 0.280 & 0.076 & 0.182 & 0.383 & 0.437 & 365.8 \\
\midrule
\multirow{2}{*}{\href{https://qwen.ai/blog?id=qwen3.5}{Qwen3.5-122B-A10B}} & FS  & 0.283 & 0.509 & 0.103 & 0.178 & 0.237 & 0.325 & 601.2 \\
 & CoT & 0.283 & 0.238 & 0.081 & 0.247 & \textbf{0.533} & \textbf{0.557} & \textbf{313.6} \\
\midrule
\multirow{2}{*}{\href{https://docs.z.ai/guides/llm/glm-5.1}{GLM-5.1}} & FS  & 0.221 & 0.495 & 0.182 & 0.260 & 0.411 & 0.487 & 503.0 \\
 & CoT & 0.288 & 0.258 & 0.049 & 0.183 & 0.411 & 0.487 & 352.6 \\
\midrule
\multirow{2}{*}{\href{https://www.minimax.io/models/text/m27}{MiniMax-M2.7}} & FS  & 0.210 & 0.471 & 0.080 & 0.135 & 0.132 & 0.228 & 587.5 \\
 & CoT & 0.185 & 0.351 & 0.061 & 0.113 & 0.140 & 0.223 & 500.9 \\
\bottomrule
\end{tabular}
\caption{Task 2 evaluation results, where the best performance is in \textbf{bold}.}
\label{tab:main-task2}
\end{table*}

Furthermore, we examine whether the evaluation is affected by same-family preference, the tendency of an LLM judge to systematically favor outputs from models of the same family \cite{NEURIPS2024_7f1f0218,chen2025llmevaluatorspreferreason}. Specifically, outputs from Qwen3.5-122B-A10B and Qwen3.5-397B-A17B, rated by safety experts, are independently evaluated by both Qwen3.6-Max and GPT-5.5. The two judges exhibit highly comparable levels of agreement with the expert ratings: across all four evaluation dimensions and three correlation measures, the maximum absolute difference between corresponding correlation coefficients is only 0.027. These results suggest that the observed judge-expert agreement is robust to the choice of evaluator and is unlikely to be driven primarily by model-family affinity.

\begin{figure}[t]
    \centering
    \includegraphics[width=\columnwidth]{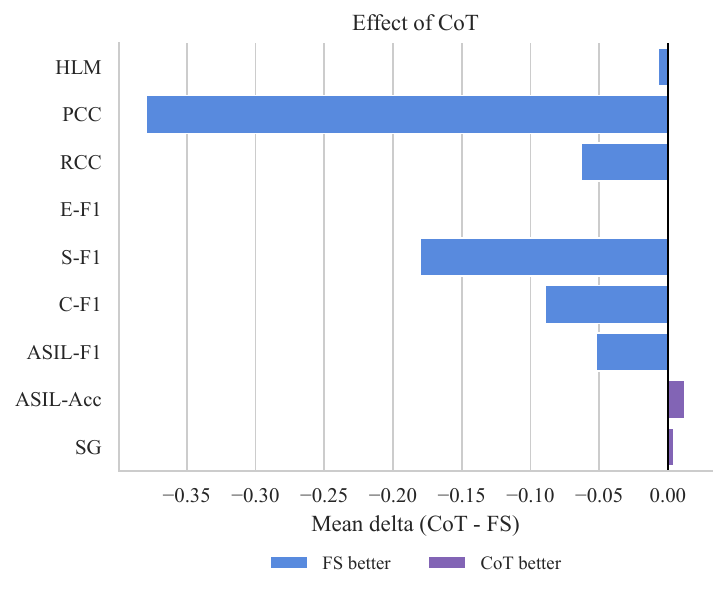}
    \caption{Average effect of CoT over few-shot prompting (FS). Positive values indicate improvement.}
    \label{fig:rq2-cot-effect}
\end{figure}

\subsection{Experimental Setup}
\label{sec:experiments:setup}

We conduct experiments on nine frontier LLMs, including GPT-5.5, GPT-5.4-mini, Gemini-3.5-Flash, DeepSeek-V4-Pro, DeepSeek-V4-Flash, Qwen3.5-397B-A17B, Qwen3.5-122B-A10B, GLM-5.1, and MiniMax-M2.7. All models are accessed through their official APIs.
Each model is evaluated under two prompting strategies: \textbf{Few-shot (FS)} provides $k{=}3$ retrieved exemplars from the training split, along with a concise schema instruction. \textbf{Chain-of-Thought (CoT)} uses the same exemplars but asks the model to externalize its causal reasoning before producing the final answer~\citep{wei2023chainofthoughtpromptingelicitsreasoning,peng2023doesincontextlearningfall}.
We keep CoT even for models with native internal reasoning because the externalized trace can help models organize their thoughts, be auditable, and support error analysis. To reduce variability, we set the temperature to 0.1 for models that support temperature configuration and retain the default decoding settings for all other models.

\subsection{Experimental Results}
\label{sec:results:overall}

Tables~\ref{tab:main-task1} and~\ref{tab:main-task2} report the evaluation results, which reveal a clear capability gap between hazard generation and risk assessment. 
We also implemented a non-LLM baseline to demonstrate the sophistication of the task beyond rule-based lookup, which is organized in Appendix \ref{app:non-llm-baseline}.
Models perform substantially better on Task~1; however, performance drops sharply on structured risk assessment, especially E/S/C prediction and downstream ASIL allocation. The best ASIL macro-F1 reaches only 0.261, showing that plausible HARA narratives do not translate into reliable standards-grounded risk classification.

\begin{figure}[t]
    \centering
    \includegraphics[width=\columnwidth]{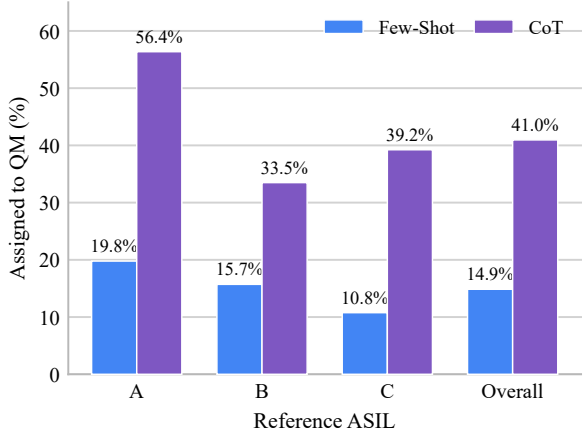}
    \caption{Rates of incorrectly assigning reference ASIL A-C cases to QM,
    pooled across nine models under few-shot prompting (FS) and CoT.}
    \label{fig:asil-to-qm-misclassification}
\end{figure}

Leveraging CoT prompting provides limited and task-dependent benefits. As summarized in Figure \ref{fig:rq2-cot-effect}, CoT generally reduces performance on HLM, PCC, Severity, Controllability, and ASIL, while consistently improving FTTI estimation. This suggests that externalized reasoning traces may help numerical timing estimation, but do not reliably improve categorical judgments governed by ISO 26262 risk semantics. 
Furthermore,  given the skewed distribution of QM in derived ASIL (see Figure \ref{fig:label-distribution}, we analyze the rates of incorrectly assigning reference ASIL A/B/C cases to QM, as illustrated in Figure \ref{fig:asil-to-qm-misclassification}. The overall false-QM rate increases by 26.1\% points under CoT, from 14.9\% to 41.0\%. For class C, the highest ASIL in SAFARI, increases from 10.8\% to 39.2\%.
Overall, current LLMs are better suited for hazard exploration and documentation support than for autonomous safety-critical risk assessment.

\subsection{Error Analysis}
\label{sec:experiments:ErrorAnalysis}

To localize the limitations of current LLMs, we manually inspect representative errors and identify two recurring failure modes that correspond to the two stages of the HARA workflow. Representative examples are provided in Appendix \ref{app:error-cases}.

\paragraph{Context Omission.}

In Task 1, models often recover the primary malfunction but fail to preserve the scenario-critical context in their descriptions. Typical omissions include road curvature, low-friction surfaces, visibility constraints, and surrounding traffic configuration, which are safety-relevant because they affect collision likelihood, driver response, and downstream risk assessment.

\paragraph{Controllability misjudgment.} In Task 2, models frequently miscalibrate the driver’s ability to avoid or mitigate the hazardous event, especially when controllability depends on interactions among vehicle dynamics, environmental conditions, traffic participants, and available reaction time. Since ASIL is deterministically derived from the predicted $(E,S,C)$ tuple, a single controllability error can propagate to a different ASIL allocation. This makes controllability assessment a key bottleneck for standards-grounded HARA reasoning.

\section{Conclusion}
\label{sec:conclusion}

We introduced SAFARI, an industrial benchmark for evaluating LLM-assisted automotive HARA under ISO~26262, covering both open-ended hazard generation and standards-grounded risk assessment, supported by a reference-anchored, expert-validated LLM-as-a-judge protocol for textual safety artifacts. Experiments on nine frontier LLMs reveal a clear gap between plausible hazard narratives and reliable ISO~26262 risk classification, and CoT prompting provides limited benefit beyond FTTI estimation. Error analysis further identifies scenario-context omission and controllability misjudgment as key bottlenecks, suggesting directions for future development.

\section*{Limitations}

We organize the limitation of SAFARI two-fold: \textbf{(1)} SAFARI focuses on a single malfunction category---unintended drive force or torque output. Although this malfunction is critical and appears across a wide range of operational scenarios, the benchmark does not cover the full diversity of automotive hazards considered in ISO 26262. Future extensions should include additional malfunction classes and vehicle domains to assess cross-failure generalization. \textbf{(2)} The benchmark is derived from industrial HARA records and therefore reflects the empirical distribution of real safety-engineering practice. How to mitigate the imbalance issue remains an open problem for future development, but it does not diminish our contribution for providing a benchmark that reflect real-world industrial distribution.

\section*{Acknowledgments}

This research was supported by the Collaborative Research Project of Xi’an Jiaotong-Liverpool University (RDS10120240248), the Fundamental Research Funds for the Central Universities, Jilin University (93K172025K19), and the Leadership Talent Program (Science and Education) of SIP, KJL2024104. This work was also supported in part by the Suzhou Industrial Park Cross-Innovation Research Platform for Affective Computing and Interactive Health (CXK2025101).


\bibliography{custom}

\appendix

\newpage

\section{Non-LLM Baselines}
\label{app:non-llm-baseline}

To highlight the sophistication of the task, we additionally implement a deterministic rule-based baseline for the E/S/C/ASIL components of Task 2. The baseline predicts E, S, and C solely from the structured scenario fields using transparent, low-order heuristics over a small number of attributes. ASIL is then derived using the same deterministic E/S/C-to-ASIL mapping applied to all model outputs.

\paragraph{Severity (S).} ISO 26262 characterizes severity according to the potential harm to persons at risk. As a conservative heuristic, we assign S3 whenever the traffic participant is a vulnerable road user, such as a pedestrian or cyclist. For other traffic participants, we use the absolute speed difference between the ego vehicle and the other agent as a coarse speed-based proxy. A difference of 30 kph or more is mapped to S3, a difference of at least 15 kph but below 30 kph to S2, a positive difference below 15 kph to S1, and no speed difference to S0.

\paragraph{Exposure (E).} ISO 26262 defines exposure as the probability of encountering an operational situation under normal or reasonably foreseeable vehicle use. We assign a base exposure level of E4 to urban-road and highway scenarios and E3 to parking scenarios. The base level is reduced by one step if either the weather is not clear or visibility is low, reflecting the lower probability of occurrence of the resulting combination of operating conditions. The exposure level is reduced only once when both conditions are present.

\paragraph{Controllability (C).} ISO 26262 considers the ability of the driver or other persons at risk to avoid or mitigate harm through timely reactions. We approximate the available reaction opportunity using driver occupancy and relative position. A scenario is assigned C2 when either no driver is present or the other agent is within the predefined safe distance; all remaining scenarios are assigned C0.

All results are computed on the same test set using the same evaluation protocol, as shown in Table \ref{tab:rule-based-baseline}. From the table, we observe that the best LLM configuration for each metric outperforms the rule-based baseline. Nevertheless, the rule-based system outperforms 12 of the 18 LLM configurations on S, 13 on C, and 5 on ASIL, while underperforming all configurations on E. These results indicate that simple scenario attributes provide a competitive signal for S and C classification, whereas the layout- and condition-based heuristics used here capture exposure poorly. More broadly, they show that improvements over a transparent non-LLM comparator are not consistent across all LLM configurations.

\begin{table}[t]
\centering
\small
\begin{tabular}{lccc}
\toprule
\multirow{2}{*}{\textbf{Dim.}}
& \multirow{2}{*}{\textbf{Rule}}
& \multirow{2}{*}{\textbf{Best LLM}}
& \textbf{\# of Configs.} \\
& & & \textbf{Outperformed} \\
\midrule
\multirow{2}{*}{E}
& \multirow{2}{*}{0.113}
& 0.318
& \multirow{2}{*}{0 / 18} \\
& & Qwen3.5-397B (FS) & \\
\midrule
\multirow{2}{*}{S}
& \multirow{2}{*}{0.444}
& 0.514
& \multirow{2}{*}{12 / 18} \\
& & GPT-5.5 (FS) & \\
\midrule
\multirow{2}{*}{C}
& \multirow{2}{*}{0.145}
& 0.216
& \multirow{2}{*}{13 / 18} \\
& & GPT-5.5 (FS) & \\
\midrule
\multirow{2}{*}{ASIL}
& \multirow{2}{*}{0.171}
& 0.261
& \multirow{2}{*}{5 / 18} \\
& & GPT-5.5 (FS) & \\
\bottomrule
\end{tabular}
\caption{Macro-F1 comparison of the rule-based baseline and the best LLM configurations.}
\label{tab:rule-based-baseline}
\end{table}

\section{Representative Error Cases}
\label{app:error-cases}

\subsection{Context Omission}
\label{app:error-context}

\paragraph{Scenario.}
\textit{urban road, ego vehicle traveling at 40 kph, lead vehicle traveling at 20 kph, with a separation greater than the safe distance (33 m).}

\paragraph{Reference Hazardous Event.}
\textit{The ego vehicle unexpectedly generates drive force while traveling on a downhill urban road and collides with a vehicle traveling 33 m ahead.}

\paragraph{Model Prediction.}
\textit{The ego vehicle unexpectedly accelerates on a straight urban road and rear-ends the vehicle ahead.}

\paragraph{Explanation.}
The model correctly identifies the malfunction, collision outcome, and affected road users, but fails to preserve critical contextual information present in the expert reference. Compared with the reference description, the prediction provides a more generic characterization of the road scenario and omits important environmental and distance-related details. Such omissions reduce the specificity of the hazardous-event formulation and weaken its alignment with expert HARA analysis.
Importantly, the missing information is not merely descriptive. In expert HARA practice, contextual factors such as road characteristics and vehicle spacing can influence the evolution of the hazard, the assessment of controllability, and ultimately the resulting risk evaluation.

\subsection{Controllability Misjudgment}
\label{app:error-control}

\paragraph{Scenario.}
\textit{Urban-road intersection, ego vehicle decelerating
at 40 kph, a vulnerable road user crossing the lane
33 m ahead.}

\paragraph{Expert Assessment.}
\textit{E4, S3, C1, ASIL B.}

\paragraph{Model Prediction.}
\textit{E4, S3, C3, ASIL D.}

\paragraph{Explanation.}
The model correctly predicts \textit{Exposure} and \textit{Severity}, but substantially overestimates the \textit{Controllability} class, assigning C3 instead of the expert label C1.
Manual inspection suggests that the model focuses primarily on collision severity while overlooking the available reaction distance and opportunities for driver intervention. In this scenario, the vulnerable road user is located 33 m ahead, and the ego vehicle is already decelerating, providing a realistic opportunity for avoidance.
Because ASIL is deterministically derived from the $(E,S,C)$ tuple, this controllability error directly propagates to the downstream risk classification, changing the final ASIL from B to D. This example illustrates how a single misjudged risk parameter can lead to a substantially different safety allocation outcome.

\end{document}